\documentclass[runningheads]{llncs}

\usepackage{eccv}

\usepackage{eccvabbrv}

\usepackage{graphicx}
\usepackage{booktabs}
\usepackage{xcolor}
\usepackage{float}
\usepackage{comment}
\usepackage{adjustbox}
\usepackage{amssymb}
\usepackage[accsupp]{axessibility}  

\usepackage{hyperref}

\usepackage{orcidlink}

\providecommand{\mathindent}{0pt}

\begin{document}
	
	\title{ODE-Based Transformer Decoders for Iterative Sign Language Translation}

    \author{
    Tuğçe Kızıltepe\inst{1,2}\orcidlink{0009-0000-6529-1871} \and
    Hacer Yalim Keles\inst{2}\orcidlink{0000-0002-1671-4126}
    }
    
    \authorrunning{T.~Kızıltepe and H.~Y.~Keles}
    
    \institute{
    ASELSAN, Ankara, Türkiye\\
    \email{tkiziltepe@aselsan.com}
    \and
    Department of Computer Engineering,
    Hacettepe University, Ankara, Türkiye\\
    \email{tugcekiziltepe@hacettepe.edu.tr}\\
    \email{hacerkeles@cs.hacettepe.edu.tr}
    }
	\maketitle

	\begin{abstract}
		Sign language translation has achieved strong results with Transformer architectures, yet recent improvements largely rely on scaling model capacity at the cost of increased computation. We propose a parameter-efficient alternative that improves expressiveness without increasing model size. Rather than scaling capacity, we focus on enhancing the update dynamics of iterative refinement decoders, where each refinement step corresponds to one internal decoder iteration that progressively improves the latent representation before translation generation. We reinterpret residual refinement updates from an Ordinary Differential Equation (ODE) perspective and replace them with higher-order numerical integration schemes, namely Runge--Kutta methods (RK-2 and RK-4). These methods perform multiple function evaluations within each refinement step to produce more accurate and stable representation updates without adding decoder parameters. To the best of our knowledge, this is the first application of ODE-inspired update dynamics to sign language translation. RK-2 achieves 22.96 BLEU-4 on the PHOENIX-14T test set and 19.34 BLEU-4 on the CSL-Daily test set, outperforming the IPSLT baseline on both benchmarks, with fewer decoder layers and refinement iterations on CSL-Daily. These results suggest that stronger refinement dynamics can improve translation performance under parameter-efficient decoder designs, providing a complementary alternative to conventional model scaling.
		
	\end{abstract}

	\section{Introduction}

\begin{figure}[!t]
    \centering
    \includegraphics[width=0.68\linewidth]{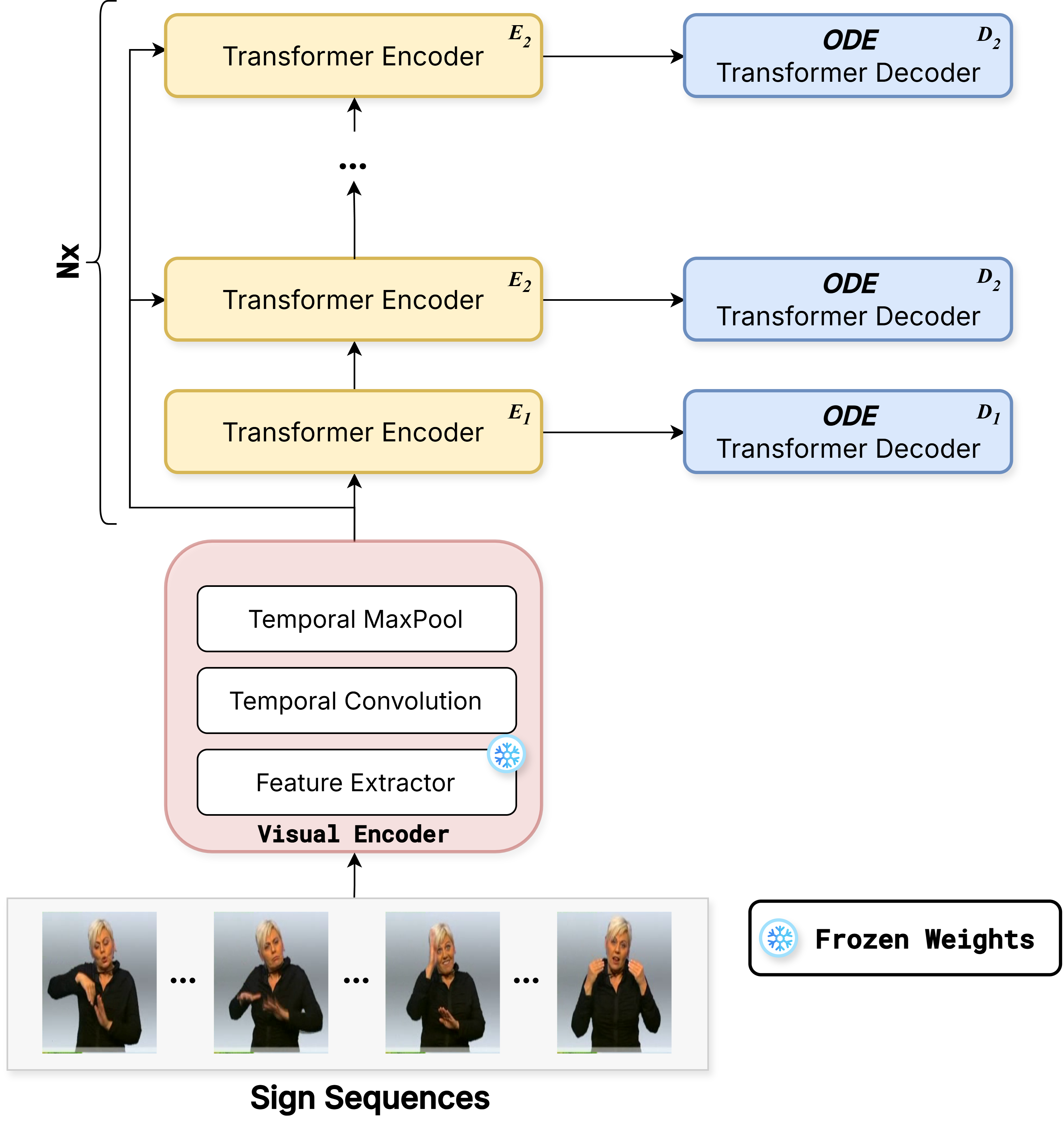}
    \caption{Overview of the proposed architecture. A frozen visual encoder
    extracts features, followed by stacked Transformer encoder layers. At each
    stage, representations are decoded by ODE-based Transformer decoders within
    an iterative refinement framework.}
    \label{fig:architecture}
\end{figure}
	Sign language translation (SLT) aims to generate fluent spoken-language text directly from continuous visual sign language input. SLT can be formulated as a gloss-based approach relying on intermediate gloss representations or as a gloss-free approach that directly maps visual input to text, where the latter offers improved scalability by eliminating costly gloss annotations. Gloss-free SLT requires jointly modeling fine-grained spatiotemporal cues, such as hand shape, motion, and facial expressions, and capturing long-range dependencies, while the lack of explicit visual--textual alignment and the linguistic differences between sign and spoken languages further complicate the task.
	
	Recent progress in SLT has been largely driven by scaling model capacity and leveraging large pretrained components, such as powerful visual encoders and large language models. While these strategies yield strong performance gains, they increase computational cost, and comparatively less attention has been given to how representations are updated across layers.
	
	Transformer-based architectures form the foundation of most modern SLT systems, where representations are updated across layers through residual connections. Such connections can be viewed as discretizations of an underlying Ordinary Differential Equation (ODE), where each layer performs a single first-order Euler integration step \cite{li-etal-2022-ode, zhong2022a}. This first-order formulation may limit the accuracy and stability of representation evolution, suggesting that more advanced numerical integration methods could improve the quality of representation updates. This viewpoint provides an alternative to scaling, where improvements arise from better update dynamics rather than larger models.
	
	In this work, we propose a parameter-efficient approach that enhances model expressiveness by improving update dynamics instead of increasing model capacity. To the best of our knowledge, this is the first work to introduce an ODE-based perspective into SLT. Building on an iterative refinement framework, we replace standard residual decoder updates with higher-order Runge–Kutta methods (RK-2 and RK-4), which approximate the underlying dynamics using multiple function evaluations per step, resulting in more accurate and stable updates.
	
	We evaluate the proposed approach on PHOENIX-14T and CSL-Daily. The results demonstrate that ODE-guided update dynamics can improve test performance without adding decoder parameters. These outcomes suggest that enhancing the quality of iterative representation updates is a promising parameter-efficient direction for SLT, complementary to conventional model scaling.

	\section{Related Work}
	
	\subsection{Sign Language Translation}

	Sign language translation (SLT) is a challenging task within the broader field of Sign Language Understanding (SLU), which aims to infer semantic meaning from visual sign language data. Earlier work in SLU has largely focused on recognition: isolated sign recognition classifies individual signs from short clips \cite{li2025uni, Sincan_2020, 11092343, zuo2023natural}, while continuous sign recognition transcribes signing video into gloss sequences \cite{10203106, li2025uni, 10205442, Wei2023ImprovingCS}, where glosses serve as intermediate representations approximating sign-level annotations.
	
	SLT can be formulated either as an end-to-end gloss-free task or as a gloss-based framework that leverages gloss supervision, either explicitly or through auxiliary alignment objectives \cite{camgoz2020signlanguagetransformersjoint, tan-etal-2025-improvement}. In explicit settings, gloss sequences are first predicted via continuous sign language recognition (CSLR) and subsequently mapped to spoken-language text \cite{zhang2023sltunet, ye-etal-2023-cross, 10.5555/3600270.3601510, Chen2022ASM, camgoz2018}. While gloss supervision provides strong intermediate constraints due to its temporal alignment with sign language, it is labor-intensive and requires domain expertise, limiting scalability. Moreover, gloss annotations are typically unavailable in large-scale open-domain datasets such as YouTube-SL25 \cite{tanzer2024youtubesl25largescaleopendomainmultilingual}, OpenASL \cite{shi2022opendomainsignlanguagetranslation}, YouTube-ASL \cite{uthus2023youtubeasllargescaleopendomainamerican}, and BOBSL \cite{albanie2021bbcoxfordbritishsignlanguage}, making gloss-free approaches a more scalable alternative for end-to-end SLT.
	
	Due to the limited size of SLT datasets and the multimodal nature of the task, transfer learning and pretraining have become key strategies. Most SLT frameworks decouple visual representation learning from language modeling, employing independently pretrained visual encoders and language decoders \cite{Chen2022ASM, Zhou_2023_ICCV, wong2024signgpt, 10.1145/3742886.3756703}. To mitigate the modality gap, recent approaches adopt visual–language pretraining with contrastive objectives and masked modeling \cite{Zhou_2023_ICCV}, while gloss-free settings introduce pseudo-gloss pretraining \cite{wong2024signgpt} and gloss-attention mechanisms \cite{yin2023gloss} to approximate gloss supervision.
	
	Recent advances increasingly leverage large pretrained and foundation models. On the language side, large language models such as GPT \cite{wong2024signgpt}, LLaMA \cite{Jang2025LostIT, Gong2024LLMsAG}, FlanT5-XL \cite{hwang-etal-2025-efficient}, and mBART \cite{Chen2024FactorizedLA} are integrated as decoders or auxiliary supervisors \cite{guo2025bridging}, enhancing linguistic fluency and long-range dependency modeling. On the visual side, foundation models pretrained on large-scale image or video corpora, including DINOv2 \cite{wong2024signgpt, gueuwou-etal-2025-signmusketeers}, CLIP \cite{Wu2025SignMouthLM, hwang-etal-2025-efficient}, and VideoMAE \cite{hwang-etal-2025-efficient}, provide strong spatiotemporal representations. While these approaches achieve substantial gains, they typically rely on increased model capacity and large pretrained backbones, reflecting a broader trend of scaling.
	
	\subsection{Transformers and Ordinary Differential Equation (ODE) Techniques}
	
	Transformer architectures have increasingly been interpreted from an ODE-based perspective, where residual layers correspond to discrete integration steps of an underlying continuous transformation. This view builds on earlier connections between Neural ODEs and equilibrium formulations of deep networks~\cite{10.5555/3327757.3327764,bai2019deep}, under which standard residual updates resemble first-order numerical solvers, motivating more expressive update rules.
	
	Recent studies extend this perspective specifically to Transformers. Zhong et al.~\cite{zhong2022a} analyze Transformer layers through a Neural ODE formulation, while Li et al.~\cite{li-etal-2022-ode} introduce Runge--Kutta-inspired updates for sequence generation, replacing the first-order residual step with higher-order integration. Tong et al.~\cite{tong2025neural} further explore continuous-depth architectures, depth-dependent parameterizations, and adaptive fine-tuning to improve flexibility and stability.
	
	Prior work has primarily applied the ODE perspective to general sequence generation or conventional Transformer layer stacks. In contrast, we study this formulation in the context of iterative SLT, where representations are repeatedly refined over multiple update steps. Building on Li et al.~\cite{li-etal-2022-ode}, we integrate Runge--Kutta updates into the refinement process.

	\subsection{Iterative Refinement}
	
	Iterative refinement is a general strategy for improving predictions by repeatedly updating intermediate representations or outputs over multiple steps. It appears across vision and language tasks in various forms, including progressive generative refinement~\cite{gregor2015drawrecurrentneuralnetwork, 8575275, ren2019progressiveimagederainingnetworks, 9711424, 9887996}, structural alignment optimization~\cite{Ling2019FastIO, Peng2020DeepSF, 9423326, Tao2022E2ECAE}, and self-feedback mechanisms~\cite{madaan2023selfrefine}.
	
	In Transformer architectures, refinement is commonly realized through recursive or weight-shared designs, where the same block is applied multiple times and the effective depth is determined by the number of iterations \cite{Bae2024RelaxedRT, xu2026loopingforwardrecursivetransformers, 10.1007/978-3-031-20053-3_42}. Related approaches, such as the Universal Transformer and depth-adaptive variants, further explore iterative computation by reusing shared layers or dynamically adjusting computation depth \cite{Dehghani2018UniversalT, Elbayad2019DepthAdaptiveT, bae2025mixtureofrecursionslearningdynamicrecursive}.
	
	In sign language translation, iterative refinement has been adopted to progressively improve decoding outputs across multiple steps~\cite{Yao2023SignLT}. Similar strategies have also been explored in sign language production, where iterative updates refine generated sign sequences~\cite{Kiziltepe_2025_ICCV}. These approaches improve what is refined or how many refinement steps are taken, while the update within each step remains a first-order residual operation. In contrast, we recast each refinement step as a higher-order numerical integration of an underlying continuous transformation, producing a more accurate per-step update without adding parameters or refinement steps, complementary to existing iterative refinement methods.
	
	\section{Methodology}
	
	We propose ODE-guided decoders for sign language translation, building on the iterative refinement framework \cite{Yao2023SignLT} and enhancing representation updates through higher-order numerical integration schemes. An overview of the proposed model architecture is illustrated in Fig.~\ref{fig:architecture}. 
	
	\subsection{Visual Features}
	
	We extract 512-dimensional frame-level visual features using dataset-specific ResNet-18 backbones pretrained under the CSLR setting. For PHOENIX-14T, we use the ResNet-18 backbone pretrained with the VAC framework \cite{Min_2021_ICCV}. For CSL-Daily, we use the ResNet-18 backbone pretrained with the Self-Emphasizing Network framework \cite{hu2023self}. To reduce training cost, the visual backbones are kept frozen during training. The extracted visual embeddings are then processed by a stack of 1D temporal convolution and max-pooling layers, which model local motion patterns and provide hierarchical temporal abstraction before the features are passed to the translation model.
	
	\subsection{Initialization}
	
	In the initialization stage, an encoder–decoder pair generates an initial prediction from visual features. The initial encoder $E_1$ processes the input features, and its representation is used to obtain the prediction $Y_0$. This representation is then passed to the iterative refinement module, where it serves as input to the subsequent encoding stage ($E_2$). The initialization decoder is used only during training and is not involved in inference.
	
	\subsection{Iterative Refinement}
	Building on the initial representation, we perform iterative refinement to progressively improve the model predictions. The refinement stage is implemented using a dedicated encoder, denoted as $E_2$, which is applied iteratively. At each iteration $k$, the model updates its representation by conditioning on both the visual features $V$ and the representation obtained from the previous iteration. Specifically, given the previous representation $H_{k-1}$, the refinement encoder $E_2$ produces an updated representation
	\begin{equation}
		\label{eq:refinement}
		\mbox{}\hfill H_k = E_2(H_{k-1}, V). \hfill\mbox{}
	\end{equation}
	
	The updated representation $H_k$ in Eq.~\ref{eq:refinement} is then passed to the decoder to generate the refined prediction $Y_k$. This process is repeated for $K$ iterations, enabling progressive improvement of the model outputs.
	
	In both the initialization and iterative refinement stages, the encoder uses the attention module introduced in~\cite{yin2023gloss} in place of standard attention blocks.
	
	\subsection{ODE-Based Transformer Decoder}
	
	To improve the quality of representation updates, we reinterpret the Transformer decoder as a dynamical system and replace standard residual updates with higher-order numerical integration schemes.
	
	In standard Transformers, each layer performs a residual update of the form:
	\begin{equation}
		\label{eq:euler}
		\mbox{}\hfill y_{t+1} = y_t + F(y_t) \hfill\mbox{}
	\end{equation}
	which can be interpreted as a first-order Euler discretization. To obtain more accurate updates, we employ higher-order Runge--Kutta methods, specifically second-order (RK-2) and fourth-order (RK-4) schemes.
	
	\paragraph{\textbf{Post-update LayerNorm.}} After computing the RK-style update at step $t$, we optionally apply Layer Normalization to the updated latent representation. For notational simplicity, we write the RK update increment using the same form as Eq.~\ref{eq:euler}, i.e., $F(y_t)$, where $F(\cdot)$ denotes the effective update.
	\[
	\hspace*{-\mathindent}\mbox{}\hfill \tilde{y}_{t+1} = y_t + F(y_t), \qquad y_{t+1} = \mathrm{LayerNorm}(\tilde{y}_{t+1}). \hfill\mbox{}
	\]
	This post-update normalization can be used to stabilize the iterative refinement process and reduce scale variation across refinement steps. 
	
	\paragraph{\textbf{Direct RK-2 Formulation (Method 1)}}
	Following prior work \cite{li-etal-2022-ode}, we adopt a direct RK-2 formulation by treating the Transformer decoder layer as the transformation function $F(\cdot)$. The update is computed as:
	\[
	\hspace*{-\mathindent}\mbox{}\hfill F_1 = F(y_t), \quad F_2 = F(y_t + F_1), \quad y_{t+1} = y_t + \tfrac{1}{2}\left(F_1 + F_2\right). \hfill\mbox{}
	\]
	This formulation directly applies RK-2 by evaluating the decoder layer multiple times, and remains compatible with standard Transformer decoder layers without architectural modifications.
	
	\paragraph{\textbf{Residual-Compatible RK-2 Formulation (Method 2)}}
	Since Transformer layers inherently follow a residual structure as shown in Eq.~\ref{eq:euler}, they can be interpreted as implementing a first-order (Euler) update, where $F(y_t)$ corresponds to the residual transformation. In practice, a Transformer decoder layer $g(\cdot)$ produces an output of the form $g(y_t) = y_t + F(y_t)$, allowing the update function to be implicitly recovered as:
	\[
	\hspace*{-\mathindent}\mbox{}\hfill F(y_t) = g(y_t) - y_t. \hfill\mbox{}
	\]
	
	Building on this observation, we derive the RK-2 update increments directly from the residual behavior of the decoder layer, without explicitly defining $F(\cdot)$. Specifically, let $g(\cdot)$ denote a Transformer decoder layer with residual connections. We compute:
	\[
	\hspace*{-\mathindent}\mbox{}\hfill \tilde{y}_1 = g(y_t), \quad F_1 = \tilde{y}_1 - y_t, \hfill\mbox{}
	\]
	\[
	\hspace*{-\mathindent}\mbox{}\hfill \tilde{y}_2 = g(y_t + F_1), \quad F_2 = \tilde{y}_2 - (y_t + F_1), \hfill\mbox{}
	\]
	and update the representation as:
	\[
	\hspace*{-\mathindent}\mbox{}\hfill y_{t+1} = y_t + \tfrac{1}{2}\left(F_1 + F_2\right). \hfill\mbox{}
	\]
	
	This formulation naturally aligns with the residual structure of Transformers, allowing RK-2 dynamics to be integrated without modifying the architecture or introducing additional parameters.
	
	\paragraph{\textbf{RK-4 Extension}}
	We further extend both formulations to the fourth-order Runge--Kutta (RK-4) scheme, which performs four intermediate evaluations within a single update step to obtain a more accurate estimate of the underlying transformation.
	
	For the direct formulation (Method 1), where $F(\cdot)$ denotes the decoder-layer transformation, the update is computed as:
	\begin{equation}
		\label{eq:rk4-direct}
		\mbox{}\hfill
		\begin{gathered}
			F_1 = F(y_t), \\
			F_2 = F\left(y_t + \tfrac{1}{2}F_1\right), \\
			F_3 = F\left(y_t + \tfrac{1}{2}F_2\right), \\
			F_4 = F\left(y_t + F_3\right), \\
			y_{t+1} = y_t + \tfrac{1}{6}\left(F_1 + 2F_2 + 2F_3 + F_4\right).
		\end{gathered}
		\hfill\mbox{}
	\end{equation}
	
	For the residual-compatible formulation (Method 2), we compute each intermediate increment from the residual behavior of the decoder layer $g(\cdot)$. That is, for each intermediate input $\hat{y}_i$, the corresponding increment is obtained as $F_i = g(\hat{y}_i) - \hat{y}_i$, where $\hat{y}_i$ follows the same RK-4 intermediate states used above. This allows the RK-4 update to be applied without removing the residual structure already present in Transformer decoder layers.
	
	In our RK-based decoder, all stage evaluations ($F_1$, $F_2$ in RK-2 and $F_1,\ldots,F_4$ in RK-4) are computed by repeated applications of the same decoder function $f_\theta$. Increasing the RK order therefore increases the number of decoder evaluations per refinement step without introducing additional decoder parameters.
	
	\subsection{Training Objective}
	
	We train the model with supervised cross-entropy and an iteration-wise distillation objective, which we refer to as the iterative distillation loss (IDL). Cross-entropy is applied to the initialization output ($E_1$) and to the final refined prediction from the refinement stage ($E_2$). During refinement, IDL is implemented as a Kullback--Leibler divergence from each intermediate refinement prediction to the final refinement prediction, where the last refinement step acts as the teacher distribution. We sum the IDL terms across refinement steps to encourage consistent refinement trajectories and reduce oscillations. 
	
	\section{Experiments}
	
	\subsection{Implementation Details}
	
	\textbf{Dataset.}
	We evaluate on PHOENIX-14T \cite{camgoz2018}, a German Sign Language benchmark derived from weather forecast broadcasts, and CSL-Daily \cite{zhou2021signbt}, a large-scale Chinese Sign Language dataset covering daily-life topics with a broader vocabulary. For both datasets, models are trained on video--translation pairs and evaluated against reference translations without ground-truth gloss supervision.
	
	\textbf{Evaluation Protocols.}
	We evaluate translation quality using BLEU \cite{papineni-etal-2002-bleu}, computed with SacreBLEU \cite{post-2018-call}, and ROUGE \cite{lin-2004-rouge}.
	
	\textbf{Training Settings.}
	The proposed model was implemented using the PyTorch \cite{pytorch} framework and trained with the AdamW \cite{Loshchilov2017DecoupledWD} optimizer using a weight decay of $1 \times 10^{-4}$. Gradient clipping with a maximum norm of 1.0 is applied to stabilize training.
	
	The learning rate follows a warm-up and cosine scheduling strategy. For PHOENIX-14T, it is warmed up from $1 \times 10^{-6}$ to $2 \times 10^{-4}$ over the first 30 epochs, with a minimum learning rate of $2 \times 10^{-6}$. For CSL-Daily, the learning rate is set to $1 \times 10^{-4}$ with a minimum learning rate of $1 \times 10^{-6}$. The batch size is set to 32, dropout is set to 0.1, and label smoothing with a factor of 0.2 is used during training.
	
	Both the encoder and decoder consist of 3 Transformer layers by default, and the feed-forward network dimension is set to 2048.
	
	\textbf{Inference.}
	During inference, we employ autoregressive decoding with beam search to generate target sequences, using only the final decoder for prediction. For both datasets, the beam width is set to 4 and the length penalty is fixed to 1.0. 
	
	\subsection{Experimental Results}
	
	\subsubsection{Comparison with SOTA}
	
	Tables~\ref{tab:sota_PHOENIX-14T} and~\ref{tab:sota_csldaily} compare our approach with recent SLT methods on PHOENIX-14T and CSL-Daily. Since existing methods differ in pretraining, visual representations, and model capacity, we include both the broader SOTA comparison and results under the same training and inference setup. Specifically, we report a non-iterative baseline with $K=0$ and reproduce IPSLT~\cite{Yao2023SignLT} using the same visual backbone. The best IPSLT setting uses $K=3$ with an IDL weight of $0.2$.
	
	For our ODE-guided decoders, we report the best configuration of each update variant selected by DEV BLEU-4, including RK2SLT-M1, RK2SLT-M2, RK4SLT-M1, and RK4SLT-M2. For the reproduced models, the tables distinguish total stored parameters from active-path training parameters (T-P) and inference parameters (I-P), and report the selected refinement count $K$, final-decoder depth, realized FLOPs, and batch-32 inference latency. On PHOENIX-14T, RK2SLT-M1 achieves the strongest BLEU-4 among our variants and improves BLEU-4 over the IPSLT baseline. On CSL-Daily, the best BLEU-4 score is achieved by RK2SLT-M2 using a single refinement iteration and a reduced decoder depth, and RK4SLT-M2 remains close behind with an even more compact single-layer decoder. Compared with recent SLT systems that often rely on larger capacity, pretrained components, or stronger visual representations, our models remain competitive by modifying only the refinement dynamics, supporting our argument that better update rules provide a parameter-efficient route to stronger translation.
	
	The test results show a dataset-dependent effect. On PHOENIX-14T, where videos are low-resolution and the domain is narrow, higher-order refinement provides more conservative gains, with RK-2 generalizing best among our variants. On the larger and more diverse CSL-Daily dataset, the benefit is clearer: RK2SLT-M2 achieves the best TEST BLEU-4, while RK4SLT-M2 is only $0.09$ BLEU-4 lower ($19.25$ vs.\ $19.34$) despite using a single refinement iteration and a single decoder layer. This indicates that RK-4 can trade additional per-step computation for reduced decoder depth and fewer refinement steps, yielding a favorable accuracy--efficiency balance.

    Tables~\ref{tab:efficiency_PHOENIX-14T} and~\ref{tab:efficiency_csldaily}
    show that the BLEU-4 gains are not obtained by increasing the active
    parameter count. On PHOENIX-14T, RK2SLT-M2 improves BLEU-4 over IPSLT by
    $0.64$ while reducing T-P, I-P, FLOPs, and measured latency by $10.6\%$,
    $12.6\%$, $14.1\%$, and $11.5\%$, respectively. RK2SLT-M1 provides a
    complementary quality-oriented configuration, improving BLEU-4 by $1.14$
    at a higher computational cost. On CSL-Daily, RK2SLT-M2 achieves the
    highest BLEU-4 of $19.34$ with $7.4\%$ fewer inference-path parameters
    than IPSLT, at modest overheads of $9.7\%$ in FLOPs and $6.3\%$ in
    measured latency. RK4SLT-M2 remains competitive at $19.25$ BLEU-4 and
    provides the most compact inference path, using $14.9\%$ fewer
    inference-path parameters than IPSLT. These results demonstrate that the
    proposed RK updates provide distinct quality--efficiency operating points
    without relying on increased model capacity.

\begin{table*}[t]
	\centering
	\caption{Comparison with state-of-the-art methods on the PHOENIX-14T TEST set.
	$^\dagger$: reproduced with our training and inference setup;
	$\sim$: approximate size; underlined values denote the best results
	among iterative methods.}
	\label{tab:sota_PHOENIX-14T}
	\begingroup
	\setlength{\tabcolsep}{4pt}
	\renewcommand{\arraystretch}{1.05}
	\scriptsize
	\begin{adjustbox}{max width=\textwidth}
		\begin{tabular}{lccc|ccccc}
			\hline
			\textbf{Method}
			& \textbf{$K$}
			& \textbf{Dec. layers}
			& \textbf{Params (M)}
			& \multicolumn{5}{c}{\textbf{TEST}} \\
			& & & & B-1 & B-2 & B-3 & B-4 & R \\
			\hline
			GFSLT-VLP~\cite{yin2023gloss}
			& -- & -- & $\sim$600
			& 43.71 & 33.18 & 26.11 & 21.44 & 42.49 \\

			SignCL~\cite{ye2024improving}
			& -- & -- & $\sim$600
			& \textbf{49.76} & \textbf{36.85} & \textbf{29.97}
			& 22.74 & \textbf{49.04} \\

			Sign2GPT with PGP~\cite{wong2024signgpt}
			& -- & -- & $\sim$1700
			& 49.54 & 35.96 & 28.83 & 22.52 & 48.90 \\

			FLa-LLM~\cite{Chen2024FactorizedLA}
			& -- & -- & $\sim$680
			& 46.29 & 35.33 & 28.03 & 23.09 & 45.27 \\

			SignLLM~\cite{Gong2024LLMsAG}
			& -- & -- & $\sim$7000
			& 45.21 & 34.78 & 28.05 & \textbf{23.40} & 44.49 \\
			\hline
			Baseline
			& 0 & 3 & 93.16
			& 43.79 & 33.92 & 27.12 & 22.54 & 45.43 \\

			IPSLT~\cite{Yao2023SignLT}$^\dagger$
			& 3 & 3 & 93.16
			& 44.09 & 33.63 & 26.48 & 21.82 & 46.35 \\

			RK2SLT-M1 (Ours)
			& 3 & 3 & 93.16
			& 44.35 & \underline{34.25} & \underline{27.53}
			& \underline{22.96} & 46.93 \\

			RK2SLT-M2 (Ours)
			& 3 & 1 & 84.75
			& \underline{44.55} & 34.21 & 27.22 & 22.46 & 46.81 \\

			RK4SLT-M1 (Ours)
			& 3 & 3 & 93.16
			& 44.19 & 34.18 & 27.48 & 22.91 & \underline{47.39} \\

			RK4SLT-M2 (Ours)
			& 3 & 2 & 88.95
			& 44.30 & 34.19 & 27.20 & 22.62 & 46.78 \\
			\hline
		\end{tabular}
	\end{adjustbox}
	\endgroup
\end{table*}

\begin{table*}[t]
\centering
\caption{Parameter counts and measured efficiency on the PHOENIX-14T TEST set.}
\label{tab:efficiency_PHOENIX-14T}
\begingroup
\setlength{\tabcolsep}{7pt}
\renewcommand{\arraystretch}{1.05}
\scriptsize

\begin{adjustbox}{max width=\textwidth}
\begin{tabular}{lcccc}
\hline
\textbf{Method}
& \textbf{T-P (M)}
& \textbf{I-P (M)}
& \textbf{FLOPs (G)}
& \textbf{Inf. ms/sample} \\
\hline

Baseline
& 54.010 & 54.010 & 54.271 & 9.015 \\

\hline

IPSLT~\cite{Yao2023SignLT}$^\dagger$
& 79.355 & 66.743 & 58.248 & 9.624 \\

RK2SLT-M2 (Ours)
& 70.947 & 58.334 & 50.012 & \textbf{8.519} \\

RK2SLT-M1 (Ours)
& 79.357 & 66.744 & 81.971 & 14.000 \\

RK4SLT-M2 (Ours)
& 75.151 & 62.538 & 104.325 & 16.967 \\

RK4SLT-M1 (Ours)
& 79.355 & 66.743 & 135.603 & 21.239 \\

\hline
\end{tabular}
\end{adjustbox}

\par\smallskip
\parbox{\textwidth}{\scriptsize
T-P and I-P denote parameters active during training and inference, respectively. Parameter counts, FLOPs, and latency exclude the visual encoder. I-P additionally excludes the initialization decoder and other training-only modules. Per-sample FP32 latency was measured on a single NVIDIA RTX 4000 Ada Generation GPU and averaged over full batches of 32 after two warm-up batches. $^\dagger$ Results reproduced using our setup.
}

\endgroup
\end{table*}
	
\begin{table*}[t]
	\centering
	\caption{Comparison with state-of-the-art methods on the CSL-Daily TEST set.
	$^\dagger$: reproduced with our training and inference setup;
	$\sim$: approximate size; underlined values denote the best results
	among iterative methods.}
	\label{tab:sota_csldaily}
	\begingroup
	\setlength{\tabcolsep}{4pt}
	\renewcommand{\arraystretch}{1.05}
	\scriptsize
	\begin{adjustbox}{max width=\textwidth}
		\begin{tabular}{lccc|ccccc}
			\hline
			\textbf{Method}
			& \textbf{$K$}
			& \textbf{Dec. layers}
			& \textbf{Params (M)}
			& \multicolumn{5}{c}{\textbf{TEST}} \\
			& & & & B-1 & B-2 & B-3 & B-4 & R \\
			\hline
			GFSLT-VLP~\cite{yin2023gloss}
			& -- & -- & $\sim$600
			& 39.37 & 24.93 & 16.26 & 11.00 & 36.44 \\

			SignCL~\cite{ye2024improving}
			& -- & -- & $\sim$600
			& 47.47 & 32.53 & 22.62 & 16.16 & 48.92 \\

			Sign2GPT with PGP~\cite{wong2024signgpt}
			& -- & -- & $\sim$1700
			& 41.75 & 28.73 & 20.60 & 15.40 & 42.36 \\

			FLa-LLM~\cite{Chen2024FactorizedLA}
			& -- & -- & $\sim$680
			& 37.13 & 25.12 & 18.38 & 14.20 & 37.25 \\

			SignLLM~\cite{Gong2024LLMsAG}
			& -- & -- & $\sim$7000
			& 39.55 & 28.13 & 20.07 & 15.75 & 39.91 \\
			\hline
			Baseline
			& 0 & 3 & 82.93
			& 43.53 & 30.94 & 22.49 & 16.91 & 44.36 \\

			IPSLT~\cite{Yao2023SignLT}$^\dagger$
			& 3 & 3 & 82.93
			& 46.22 & 33.37 & 24.52 & 18.48 & 47.07 \\

			RK2SLT-M1 (Ours)
			& 2 & 3 & 82.93
			& 48.38 & 34.88 & 25.40 & 18.95 & 48.21 \\

			RK2SLT-M2 (Ours)
			& 1 & 2 & 78.73
			& 48.02 & 34.80 & 25.60
			& \textbf{\underline{19.34}} & 48.69 \\

			RK4SLT-M1 (Ours)
			& 3 & 3 & 82.93
			& 46.62 & 33.64 & 24.68 & 18.56 & 46.91 \\

			RK4SLT-M2 (Ours)
			& 1 & 1 & 74.53
			& \textbf{\underline{48.94}}
			& \textbf{\underline{35.33}}
			& \textbf{\underline{25.79}}
			& 19.25
			& \textbf{\underline{49.37}} \\
			\hline
		\end{tabular}
	\end{adjustbox}
	\endgroup
\end{table*}

\begin{table*}[t]
\centering
\caption{Parameter counts and measured efficiency on the CSL-Daily TEST set.}
\label{tab:efficiency_csldaily}
\begingroup
\setlength{\tabcolsep}{7pt}
\renewcommand{\arraystretch}{1.05}
\scriptsize
\begin{adjustbox}{max width=\textwidth}
\begin{tabular}{lcccc}
\hline
\textbf{Method}
& \textbf{T-P (M)}
& \textbf{I-P (M)}
& \textbf{FLOPs (G)}
& \textbf{Inf. ms/sample} \\
\hline
Baseline
& 43.787 & 43.787 & 14.260
& 6.887 \\
\hline
IPSLT~\cite{Yao2023SignLT}$^\dagger$
& 69.131 & 56.519 & 18.851
& \textbf{7.901} \\

RK2SLT-M2 (Ours)
& 64.927 & 52.315 & 20.671
& 8.396 \\

RK4SLT-M2 (Ours)
& 60.724 & 48.112 & 20.729
& 8.587 \\

RK2SLT-M1 (Ours)
& 69.133 & 56.520 & 29.069
& 11.388 \\

RK4SLT-M1 (Ours)
& 69.132 & 56.520 & 51.430
& 18.900 \\
\hline
\end{tabular}
\end{adjustbox}

\par\smallskip
\parbox{\textwidth}{\scriptsize
T-P and I-P denote parameters active during training and inference, respectively. Parameter counts, FLOPs, and latency exclude the visual encoder. I-P additionally excludes the initialization decoder and other training-only modules. Per-sample FP32 latency was measured on a single NVIDIA RTX 4000 Ada Generation GPU and averaged over full batches of 32 after two warm-up batches. $^\dagger$ Results reproduced using our setup.
}
\endgroup
\end{table*}

	\subsubsection{Ablation Studies}
	\begingroup
	\setlength{\intextsep}{4pt}
	\setlength{\textfloatsep}{6pt}
	\setlength{\floatsep}{4pt}
	\setlength{\abovecaptionskip}{2pt}
	\setlength{\belowcaptionskip}{2pt}
	
	We conduct structured sequential ablations on the DEV sets of PHOENIX-14T and CSL-Daily. Each table isolates one design factor at a time, including coefficients, ODE-decoder placement, post-update LayerNorm, iterative distillation loss (IDL) weight, refinement iteration count, and decoder depth; after each block, the setting with the best BLEU-4 is selected and fixed for the following blocks. Following the IPSLT baseline setting, we report configurations with up to three refinement iterations and three decoder layers, avoiding larger configurations to preserve a controlled parameter-efficient comparison. We denote the RK-2 variants as RK2SLT-M1 and RK2SLT-M2, and the RK-4 variants as RK4SLT-M1 and RK4SLT-M2, where M1 and M2 refer to the corresponding update methods.
	
	\paragraph{PHOENIX-14T.}
	The PHOENIX-14T ablations are shown in Tables~\ref{tab:phoenix_rk2_m1_structured_ablation}--\ref{tab:phoenix_rk4_m2_structured_ablation}. For RK-2, we first ablate the coefficient formulation, while RK-4 uses fixed fourth-order coefficients and therefore starts from the placement and normalization block.
	
	\paragraph{RK2SLT-M1.}
	Table~\ref{tab:phoenix_rk2_m1_structured_ablation} selects the learnable RK-2 combination $\alpha F_1+(1-\alpha)F_2$; the averaged update $\frac{1}{2}(F_1+F_2)$ yields slightly higher ROUGE but lower BLEU-4. The best configuration applies the ODE update to all decoder layers without post-update LayerNorm, uses IDL weight $0.3$, and retains three refinement iterations and three decoder layers, indicating that this variant benefits from full refinement capacity on PHOENIX-14T.
	
	\paragraph{RK2SLT-M2.}
	Table~\ref{tab:phoenix_rk2_m2_structured_ablation} favors the averaged update $\frac{1}{2}(F_1+F_2)$, only marginally ahead of the learnable combination, suggesting Method~2 is less sensitive to coefficient weighting. It selects refinement-only placement without LayerNorm, IDL weight $0.3$, and three refinement iterations, while the depth ablation prefers a single decoder layer.
	
	\begin{table}[!htb]
		\centering
		\caption{Structured sequential ablations for RK2SLT-M1 and RK2SLT-M2 on the PHOENIX-14T DEV set. BLEU-4 is the selection criterion; within each block, the selected setting of each variant is marked with $\checkmark$ and fixed for subsequent blocks.}
		\label{tab:phoenix_rk2_m1_structured_ablation}
		\label{tab:phoenix_rk2_m2_structured_ablation}
		\begingroup
		\scriptsize
		\setlength{\tabcolsep}{2pt}
		\renewcommand{\arraystretch}{0.86}
		\setlength{\aboverulesep}{0.15ex}
		\setlength{\belowrulesep}{0.15ex}
		\begin{tabular}{@{}l@{\hspace{0.8em}}c@{\hspace{0.45em}}c@{\hspace{0.75em}}c@{\hspace{1.6em}}c@{\hspace{0.45em}}c@{\hspace{0.75em}}c@{}}
			\toprule
			\rule{0pt}{2.65ex} & \multicolumn{3}{c}{\textbf{RK2SLT-M1}} & \multicolumn{3}{c}{\textbf{RK2SLT-M2}} \\
			\cmidrule(lr){2-4}\cmidrule(lr){5-7}
			\rule{0pt}{2.65ex}\textbf{Configuration} & & \textbf{BLEU-4} & \textbf{ROUGE} & & \textbf{BLEU-4} & \textbf{ROUGE} \\
			\midrule
			\multicolumn{7}{@{}l}{\textit{Coefficients}} \\
			\midrule
			$F_1+F_2$ &  & 23.17 & 48.23 &  & 23.02 & 47.62  \\
			$\frac{1}{2}(F_1+F_2)$ &  & 23.90 & \textbf{48.96} & $\checkmark$ & \textbf{23.54} & \textbf{47.93}  \\
			$\alpha F_1+(1-\alpha)F_2$ & $\checkmark$ & \textbf{24.06} & 48.61 &  & 23.52 & 47.92  \\
			\midrule
			\multicolumn{7}{@{}l}{\textit{Placement}} \\
			\midrule
			Refinement layers &  & 23.30 & 47.79 & $\checkmark$ & \textbf{23.92} & \textbf{48.70}  \\
			Refinement layers + LN &  & 23.71 & 47.99 &  & 23.83 & 48.31  \\
			All layers & $\checkmark$ & \textbf{24.06} & \textbf{48.61} &  & 23.54 & 47.93  \\
			All layers + LN &  & 23.56 & 48.18 &  & 23.47 & 48.19  \\
			\midrule
			\multicolumn{7}{@{}l}{\textit{IDL Weight}} \\
			\midrule
			0.2 &  & 23.58 & 48.48 &  & 23.63 & 47.96  \\
			0.3 & $\checkmark$ & \textbf{24.06} & 48.61 & $\checkmark$ & \textbf{23.92} & \textbf{48.70}  \\
			0.4 &  & 23.96 & \textbf{48.98} &  & 23.50 & 48.06  \\
			0.5 &  & 23.58 & 47.27 &  & 23.36 & 48.28  \\
			\midrule
			\multicolumn{7}{@{}l}{\textit{Refinement Iterations $K$}} \\
			\midrule
			1 &  & 23.56 & 48.40 &  & 23.33 & 48.06  \\
			2 &  & 23.39 & 47.93 &  & 22.91 & 47.46  \\
			3 & $\checkmark$ & \textbf{24.06} & \textbf{48.61} & $\checkmark$ & \textbf{23.92} & \textbf{48.70}  \\
			\midrule
			\multicolumn{7}{@{}l}{\textit{Decoder Depth}} \\
			\midrule
			1 &  & 22.90 & 47.22 & $\checkmark$ & \textbf{23.99} & 48.37  \\
			2 &  & 23.55 & 47.65 &  & 23.72 & 48.44  \\
			3 & $\checkmark$ & \textbf{24.06} & \textbf{48.61} &  & 23.92 & \textbf{48.70}  \\
			\bottomrule
		\end{tabular}
		\endgroup
	\end{table}
	
	\paragraph{RK4SLT-M1.}
	Table~\ref{tab:phoenix_rk4_m1_structured_ablation} selects the RK-4 update on all decoder layers without LayerNorm. IDL weight $0.2$ is chosen, and BLEU-4 favors three refinement iterations and three decoder layers.
	
	\paragraph{RK4SLT-M2.}
	
	As shown in Table~\ref{tab:phoenix_rk4_m2_structured_ablation}, the strongest RK4SLT-M2 configuration applies the RK-4 update only within the refinement stage without the additional LayerNorm, in line with the broader trend that LayerNorm degrades BLEU-4. The best configuration uses IDL weight $0.3$ and three refinement iterations, while a two-layer decoder outperforms the three-layer one.
	
	\begin{table}[!htb]
		\centering
		\caption{Structured sequential ablations for RK4SLT-M1 and RK4SLT-M2 on the PHOENIX-14T DEV set. BLEU-4 is the selection criterion; within each block, the selected setting of each variant is marked with $\checkmark$ and fixed for subsequent blocks. RK-4 coefficients are fixed.}
		\label{tab:phoenix_rk4_m1_structured_ablation}
		\label{tab:phoenix_rk4_m2_structured_ablation}
		\begingroup
		\scriptsize
		\setlength{\tabcolsep}{2pt}
		\renewcommand{\arraystretch}{0.86}
		\setlength{\aboverulesep}{0.15ex}
		\setlength{\belowrulesep}{0.15ex}
		\begin{tabular}{@{}l@{\hspace{0.8em}}c@{\hspace{0.45em}}c@{\hspace{0.75em}}c@{\hspace{1.6em}}c@{\hspace{0.45em}}c@{\hspace{0.75em}}c@{}}
			\toprule
			\rule{0pt}{2.65ex} & \multicolumn{3}{c}{\textbf{RK4SLT-M1}} & \multicolumn{3}{c}{\textbf{RK4SLT-M2}} \\
			\cmidrule(lr){2-4}\cmidrule(lr){5-7}
			\rule{0pt}{2.65ex}\textbf{Configuration} & & \textbf{BLEU-4} & \textbf{ROUGE} & & \textbf{BLEU-4} & \textbf{ROUGE} \\
			\midrule
			\multicolumn{7}{@{}l}{\textit{Placement}} \\
			\midrule
			Refinement layers &  & 23.53 & 47.67 & $\checkmark$ & \textbf{23.71} & \textbf{48.29}  \\
			Refinement layers + LN &  & 23.37 & 47.88 &  & 23.43 & 47.39  \\
			All layers & $\checkmark$ & \textbf{23.78} & 48.14 &  & 23.66 & 48.02  \\
			All layers + LN &  & 23.75 & \textbf{48.63} &  & 23.42 & 48.11  \\
			\midrule
			\multicolumn{7}{@{}l}{\textit{IDL Weight}} \\
			\midrule
			0.1 &  & 23.74 & \textbf{48.57} &  & 23.02 & 47.14  \\
			0.2 & $\checkmark$ & \textbf{23.90} & 48.03 &  & 23.59 & 47.53  \\
			0.3 &  & 23.78 & 48.14 & $\checkmark$ & \textbf{23.71} & \textbf{48.29}  \\
			0.4 &  & 23.53 & 47.60 &  & 23.14 & 47.42  \\
			\midrule
			\multicolumn{7}{@{}l}{\textit{Refinement Iterations $K$}} \\
			\midrule
			1 &  & 23.80 & \textbf{49.12} &  & 23.16 & \textbf{48.54}  \\
			2 &  & 23.82 & 48.65 &  & 23.25 & 47.44  \\
			3 & $\checkmark$ & \textbf{23.90} & 48.03 & $\checkmark$ & \textbf{23.71} & 48.29  \\
			\midrule
			\multicolumn{7}{@{}l}{\textit{Decoder Depth}} \\
			\midrule
			1 &  & 23.62 & \textbf{48.52} &  & 22.73 & 47.46  \\
			2 &  & 23.64 & 48.03 & $\checkmark$ & \textbf{23.92} & \textbf{48.69}  \\
			3 & $\checkmark$ & \textbf{23.90} & 48.03 &  & 23.71 & 48.29  \\
			\bottomrule
		\end{tabular}
		\endgroup
	\end{table}
	
	\paragraph{CSL-Daily.}
	The CSL-Daily ablations are reported in Tables~\ref{tab:csldaily_rk2_m1_structured_ablation}--\ref{tab:csldaily_rk4_m2_structured_ablation}. Compared with PHOENIX-14T, the selected settings more often favor fewer refinement iterations or shallower decoders, indicating that the most effective refinement dynamics are dataset- and method-dependent.
	
	\paragraph{RK2SLT-M1.}
	Table~\ref{tab:csldaily_rk2_m1_structured_ablation} selects the averaged RK-2 update, refinement-only placement with LayerNorm, and a small IDL weight of $0.05$, consistent with CSL-Daily favoring lighter intermediate supervision. Two refinement iterations outperform one and three, and three decoder layers are selected, though the depth trade-off is relatively flat.
	
	\paragraph{RK2SLT-M2.}
	Table~\ref{tab:csldaily_rk2_m2_structured_ablation} selects the averaged RK-2 update over the unscaled and learnable alternatives, refinement-only placement without LayerNorm, IDL weight $0.1$, a single refinement iteration, and two decoder layers. This compact configuration also gives the strongest TEST BLEU-4 among our variants on CSL-Daily.
	
	\begin{table}[!htb]
		\centering
		\caption{Structured sequential ablations for RK2SLT-M1 and RK2SLT-M2 on the CSL-Daily DEV set. BLEU-4 is the selection criterion; within each block, the selected setting of each variant is marked with $\checkmark$ and fixed for subsequent blocks.}
		\label{tab:csldaily_rk2_m1_structured_ablation}
		\label{tab:csldaily_rk2_m2_structured_ablation}
		\begingroup
		\scriptsize
		\setlength{\tabcolsep}{2pt}
		\renewcommand{\arraystretch}{0.86}
		\setlength{\aboverulesep}{0.15ex}
		\setlength{\belowrulesep}{0.15ex}
		\begin{tabular}{@{}l@{\hspace{0.8em}}c@{\hspace{0.45em}}c@{\hspace{0.75em}}c@{\hspace{1.6em}}c@{\hspace{0.45em}}c@{\hspace{0.75em}}c@{}}
			\toprule
			\rule{0pt}{2.65ex} & \multicolumn{3}{c}{\textbf{RK2SLT-M1}} & \multicolumn{3}{c}{\textbf{RK2SLT-M2}} \\
			\cmidrule(lr){2-4}\cmidrule(lr){5-7}
			\rule{0pt}{2.65ex}\textbf{Configuration} & & \textbf{BLEU-4} & \textbf{ROUGE} & & \textbf{BLEU-4} & \textbf{ROUGE} \\
			\midrule
			\multicolumn{7}{@{}l}{\textit{Coefficients}} \\
			\midrule
			$F_1+F_2$ &  & 18.47 & \textbf{47.61} &  & 17.14 & 44.94  \\
			$\frac{1}{2}(F_1+F_2)$ & $\checkmark$ & \textbf{18.67} & 47.50 & $\checkmark$ & \textbf{18.19} & \textbf{46.50}  \\
			$\alpha F_1+(1-\alpha)F_2$ &  & 18.59 & 46.91 &  & 17.62 & 45.94  \\
			\midrule
			\multicolumn{7}{@{}l}{\textit{Placement}} \\
			\midrule
			Refinement layers &  & 17.97 & 46.46 & $\checkmark$ & \textbf{18.19} & \textbf{46.50}  \\
			Refinement layers + LN & $\checkmark$ & \textbf{18.67} & \textbf{47.50} &  & 17.76 & 46.06  \\
			All layers &  & 18.11 & 46.96 &  & 18.14 & 46.33  \\
			All layers + LN &  & 18.66 & 47.29 &  & 17.79 & 46.01  \\
			\midrule
			\multicolumn{7}{@{}l}{\textit{IDL Weight}} \\
			\midrule
			0.05 & $\checkmark$ & \textbf{18.84} & \textbf{47.88} &  & 17.84 & 46.08  \\
			0.1 &  & 18.67 & 47.29 & $\checkmark$ & \textbf{18.19} & \textbf{46.50}  \\
			0.2 &  & 18.67 & 46.93 &  & 18.00 & 46.14  \\
			0.3 &  & 18.67 & 47.29 &  & 17.81 & 46.33  \\
			\midrule
			\multicolumn{7}{@{}l}{\textit{Refinement Iterations $K$}} \\
			\midrule
			1 &  & 18.86 & 47.84 & $\checkmark$ & \textbf{18.49} & \textbf{47.58}  \\
			2 & $\checkmark$ & \textbf{19.17} & \textbf{48.11} &  & 18.02 & 46.73  \\
			3 &  & 18.84 & 47.88 &  & 18.19 & 46.50  \\
			\midrule
			\multicolumn{7}{@{}l}{\textit{Decoder Depth}} \\
			\midrule
			1 &  & 18.90 & \textbf{48.43} &  & 18.72 & 47.59  \\
			2 &  & 19.07 & 48.03 & $\checkmark$ & \textbf{18.96} & \textbf{47.78}  \\
			3 & $\checkmark$ & \textbf{19.17} & 48.11 &  & 18.49 & 47.58  \\
			\bottomrule
		\end{tabular}
		\endgroup
	\end{table}
	
	\paragraph{RK4SLT-M1.}
	Table~\ref{tab:csldaily_rk4_m1_structured_ablation} selects the RK-4 update in the refinement decoder layers with LayerNorm. IDL weight $0.2$ is chosen, and BLEU-4 favors three refinement iterations and three decoder layers. RK4SLT-M1 improves longer n-gram consistency through deeper refinement.
	
	\paragraph{RK4SLT-M2.}
	Table~\ref{tab:csldaily_rk4_m2_structured_ablation} shows the strongest efficiency-oriented trend: refinement-only placement with LayerNorm is best, while all-layer LayerNorm degrades performance sharply. With IDL weight $0.3$, both the iteration and depth blocks select the single-step, single-layer setting, so RK4SLT-M2 exploits the stronger RK-4 update in a compact configuration.
	
	\begin{table}[!htb]
		\centering
		\caption{Structured sequential ablations for RK4SLT-M1 and RK4SLT-M2 on the CSL-Daily DEV set. BLEU-4 is the selection criterion; within each block, the selected setting of each variant is marked with $\checkmark$ and fixed for subsequent blocks. RK-4 coefficients are fixed.}
		\label{tab:csldaily_rk4_m1_structured_ablation}
		\label{tab:csldaily_rk4_m2_structured_ablation}
		\begingroup
		\scriptsize
		\setlength{\tabcolsep}{2pt}
		\renewcommand{\arraystretch}{0.86}
		\setlength{\aboverulesep}{0.15ex}
		\setlength{\belowrulesep}{0.15ex}
		\begin{tabular}{@{}l@{\hspace{0.8em}}c@{\hspace{0.45em}}c@{\hspace{0.75em}}c@{\hspace{1.6em}}c@{\hspace{0.45em}}c@{\hspace{0.75em}}c@{}}
			\toprule
			\rule{0pt}{2.65ex} & \multicolumn{3}{c}{\textbf{RK4SLT-M1}} & \multicolumn{3}{c}{\textbf{RK4SLT-M2}} \\
			\cmidrule(lr){2-4}\cmidrule(lr){5-7}
			\rule{0pt}{2.65ex}\textbf{Configuration} & & \textbf{BLEU-4} & \textbf{ROUGE} & & \textbf{BLEU-4} & \textbf{ROUGE} \\
			\midrule
			\multicolumn{7}{@{}l}{\textit{Placement}} \\
			\midrule
			Refinement layers &  & 17.79 & 46.30 &  & 17.98 & 45.87  \\
			Refinement layers + LN & $\checkmark$ & \textbf{18.68} & \textbf{46.85} & $\checkmark$ & \textbf{18.03} & \textbf{46.05}  \\
			All layers &  & 17.95 & 46.13 &  & 17.95 & 45.95  \\
			All layers + LN &  & 18.26 & 46.70 &  & 16.73 & 43.83  \\
			\midrule
			\multicolumn{7}{@{}l}{\textit{IDL Weight}} \\
			\midrule
			0.1 &  & 18.48 & \textbf{47.03} &  & 17.91 & \textbf{46.55}  \\
			0.2 & $\checkmark$ & \textbf{18.68} & 46.85 &  & 18.03 & 46.05  \\
			0.3 &  & 18.33 & 46.69 & $\checkmark$ & \textbf{18.04} & 45.91  \\
			0.4 &  & 18.03 & 46.39 &  & 17.63 & 45.99  \\
			\midrule
			\multicolumn{7}{@{}l}{\textit{Refinement Iterations $K$}} \\
			\midrule
			1 &  & 18.58 & \textbf{48.10} & $\checkmark$ & \textbf{18.51} & \textbf{47.34}  \\
			2 &  & 18.38 & 46.35 &  & 17.93 & 45.97  \\
			3 & $\checkmark$ & \textbf{18.68} & 46.85 &  & 18.04 & 45.91  \\
			\midrule
			\multicolumn{7}{@{}l}{\textit{Decoder Depth}} \\
			\midrule
			1 &  & 18.21 & \textbf{47.57} & $\checkmark$ & \textbf{18.76} & \textbf{48.33}  \\
			2 &  & 18.18 & 46.94 &  & 18.34 & 48.11  \\
			3 & $\checkmark$ & \textbf{18.68} & 46.85 &  & 18.51 & 47.34  \\
			\bottomrule
		\end{tabular}
		\endgroup
	\end{table}
	
	\paragraph{Overall observations.}
	Across both datasets, the optimal ODE-decoder design is method- and dataset-dependent. For RK-2, coefficient scaling matters, with averaged or learnable combinations outperforming the unscaled update. ODE-update placement and post-update LayerNorm are not universally beneficial, and IDL requires moderate weighting, as stronger supervision can reduce final BLEU-4. Finally, PHOENIX-14T generally favors more refinement iterations, whereas CSL-Daily admits compact configurations with fewer iterations and shallower decoders, supporting the accuracy--efficiency trade-off of ODE-guided refinement.
	\endgroup
	
	\section{Conclusion}
	
	In this work, we introduced ODE-guided Transformer decoders for iterative sign language translation, replacing standard residual updates with Runge--Kutta-inspired formulations that improve decoder update dynamics without adding model parameters. Experiments on PHOENIX-14T and CSL-Daily show that the proposed updates improve TEST BLEU-4 over the matched IPSLT baseline, while the ablation studies identify dataset-specific choices for coefficient weighting, placement, normalization, IDL weighting, and refinement depth. On CSL-Daily, the fourth-order RK4SLT-M2 is especially notable: it stays within $0.09$ TEST BLEU-4 of the best variant while using a single refinement iteration and a single decoder layer, so its extra per-step function evaluations effectively substitute for decoder depth and repeated refinement. Since RK-2 and RK-4 involve multiple function evaluations, we distinguish parameter efficiency from computational efficiency and report iteration counts and decoder depth to clarify the accuracy--efficiency trade-off. These results suggest that refining how representations are updated is a promising direction, complementary to scaling pretrained models for SLT.

	\section*{Acknowledgments}
This work is supported by the Scientific and Technological Research Council of T\"urkiye (T\"UB\.{I}TAK) under the 1001 Scientific and Technological Research Projects Funding Program (Project No.~124E618). We acknowledge the EuroHPC Joint Undertaking for awarding us access to Vega at IZUM, Slovenia, through Development Access allocation 2025D08-090.

	
	%
	%
	\bibliographystyle{splncs04}
	\bibliography{main}

@String(CVPR  = {IEEE Conf. Comput. Vis. Pattern Recog.})

@String(ICCV  = {Int. Conf. Comput. Vis.})

@String(ECCV  = {Eur. Conf. Comput. Vis.})

@String(NeurIPS = {Adv. Neural Inform. Process. Syst.})

@String(ICLR  = {Int. Conf. Learn. Represent.})

@String(CVPRW = {IEEE Conf. Comput. Vis. Pattern Recog. Worksh.})

@String(AAAI  = {AAAI})

@String(ICASSP=	{ICASSP})

@String(CVPR  = {CVPR})

@String(ICCV  = {ICCV})

@String(ECCV  = {ECCV})

@String(NeurIPS = {NeurIPS})

@String(ICLR  = {ICLR})

@String(CVPRW = {CVPRW})

@INPROCEEDINGS{camgoz2018,
  author={Camgoz, Necati Cihan and Hadfield, Simon and Koller, Oscar and Ney, Hermann and Bowden, Richard},
  booktitle={2018 IEEE/CVF Conference on Computer Vision and Pattern Recognition}, 
  title={Neural Sign Language Translation}, 
  year={2018},
  volume={},
  number={},
  pages={7784-7793},
  doi={10.1109/CVPR.2018.00812}}

@inproceedings{zhou2021signbt,
  title     = {Improving Sign Language Translation with Monolingual Data by Sign Back-Translation},
  author    = {Hao Zhou and Wengang Zhou and Weizhen Qi and Junfu Pu and Houqiang Li},
  booktitle = {Proceedings of the IEEE/CVF Conference on Computer Vision and Pattern Recognition (CVPR)},
  year      = {2021},
  pages     = {1316--1325}
}

@inproceedings{papineni-etal-2002-bleu,
    title = "{B}leu: a Method for Automatic Evaluation of Machine Translation",
    author = "Papineni, Kishore  and
      Roukos, Salim  and
      Ward, Todd  and
      Zhu, Wei-Jing",
    editor = "Isabelle, Pierre  and
      Charniak, Eugene  and
      Lin, Dekang",
    booktitle = "Proceedings of the 40th Annual Meeting of the Association for Computational Linguistics",
    month = jul,
    year = "2002",
    address = "Philadelphia, Pennsylvania, USA",
    publisher = "Association for Computational Linguistics",
    doi = "10.3115/1073083.1073135",
    pages = "311--318"
}

@inproceedings{lin-2004-rouge,
    title = "{ROUGE}: A Package for Automatic Evaluation of Summaries",
    author = "Lin, Chin-Yew",
    booktitle = "Text Summarization Branches Out",
    month = jul,
    year = "2004",
    address = "Barcelona, Spain",
    publisher = "Association for Computational Linguistics",
    url = "https://aclanthology.org/W04-1013/",
    note = {Accessed 8 July 2026},
    pages = "74--81"
}

@InProceedings{Min_2021_ICCV,
    author    = {Min, Yuecong and Hao, Aiming and Chai, Xiujuan and Chen, Xilin},
    title     = {Visual Alignment Constraint for Continuous Sign Language Recognition},
    booktitle = {Proceedings of the IEEE/CVF International Conference on Computer Vision (ICCV)},
    month     = {October},
    year      = {2021},
    pages     = {11542-11551}
}

@inproceedings{Loshchilov2017DecoupledWD,
  title={Decoupled Weight Decay Regularization},
  author={Ilya Loshchilov and Frank Hutter},
  booktitle={International Conference on Learning Representations (ICLR)},
  year={2019},
  url={https://openreview.net/forum?id=Bkg6RiCqY7},
}

@inproceedings{li2025uni,
  title={Uni-Sign: Toward Unified Sign Language Understanding at Scale},
  author={Li, Zecheng and Zhou, Wengang and Zhao, Weichao and Wu, Kepeng and Hu, Hezhen and Li, Houqiang},
  booktitle={International Conference on Learning Representations (ICLR)},
  year={2025},
  url={https://proceedings.iclr.cc/paper_files/paper/2025/hash/260a14acce2a89dad36adc8eefe7c59e-Abstract-Conference.html}
}

@article{Sincan_2020,
   title={AUTSL: A Large Scale Multi-Modal Turkish Sign Language Dataset and Baseline Methods},
   volume={8},
   ISSN={2169-3536},
   DOI={10.1109/access.2020.3028072},
   journal={IEEE Access},
   publisher={Institute of Electrical and Electronics Engineers (IEEE)},
   author={Sincan, Ozge Mercanoglu and Keles, Hacer Yalim},
   year={2020},
   pages={181340–181355} }

@inproceedings{pytorch,
  title={PyTorch: An Imperative Style, High-Performance Deep Learning Library},
  author={Paszke, Adam and Gross, Sam and Massa, Francisco and Lerer, Adam and Bradbury, James and Chanan, Gregory and Killeen, Trevor and Lin, Zeming and Gimelshein, Natalia and Antiga, Luca and Desmaison, Alban and Köpf, Andreas and Yang, Edward and DeVito, Zachary and Raison, Martin and Tejani, Alykhan and Chilamkurthy, Sasank and Steiner, Benoit and Fang, Lu and Bai, Junjie and Chintala, Soumith},
  booktitle={Advances in Neural Information Processing Systems},
  volume={32},
  year={2019},
  url={https://proceedings.neurips.cc/paper/2019/hash/bdbca288fee7f92f2bfa9f7012727740-Abstract.html},
}

@INPROCEEDINGS{11092343,
  author={Li, Yuhao and Chen, Xinyue and Li, Hongkai and Pu, Xiaorong and Jin, Peng and Ren, Yazhou},
  booktitle={2025 IEEE/CVF Conference on Computer Vision and Pattern Recognition (CVPR)}, 
  title={VSNet: Focusing on the Linguistic Characteristics of Sign Language}, 
  year={2025},
  volume={},
  number={},
  pages={24320-24330},
  doi={10.1109/CVPR52734.2025.02265}}

@inproceedings{zuo2023natural,
  title={Natural Language-Assisted Sign Language Recognition},
  author={Zuo, Ronglai and Wei, Fangyun and Mak, Brian},
  booktitle={Proceedings of the IEEE/CVF Conference on Computer Vision and Pattern Recognition (CVPR)},
  pages={14890--14900},
  year={2023}
}

@INPROCEEDINGS{10203106,
  author={Guo, Leming and Xue, Wanli and Guo, Qing and Liu, Bo and Zhang, Kaihua and Yuan, Tiantian and Chen, Shengyong},
  booktitle={2023 IEEE/CVF Conference on Computer Vision and Pattern Recognition (CVPR)}, 
  title={Distilling Cross-Temporal Contexts for Continuous Sign Language Recognition}, 
  year={2023},
  volume={},
  number={},
  pages={10771-10780},
  doi={10.1109/CVPR52729.2023.01037}}

@INPROCEEDINGS{10205442,
  author={Hu, Lianyu and Gao, Liqing and Liu, Zekang and Feng, Wei},
  booktitle={2023 IEEE/CVF Conference on Computer Vision and Pattern Recognition (CVPR)}, 
  title={Continuous Sign Language Recognition with Correlation Network}, 
  year={2023},
  volume={},
  number={},
  pages={2529-2539},
  doi={10.1109/CVPR52729.2023.00249}}

@inproceedings{Wei2023ImprovingCS,
  title={Improving Continuous Sign Language Recognition with Cross-Lingual Signs},
  author={Wei, Fangyun and Chen, Yutong},
  booktitle={Proceedings of the IEEE/CVF International Conference on Computer Vision (ICCV)},
  year={2023},
  pages={23612--23621},
}

@inproceedings{camgoz2020signlanguagetransformersjoint,
  title={Sign Language Transformers: Joint End-to-End Sign Language Recognition and Translation},
  author={Camgoz, Necati Cihan and Koller, Oscar and Hadfield, Simon and Bowden, Richard},
  booktitle={Proceedings of the IEEE/CVF Conference on Computer Vision and Pattern Recognition (CVPR)},
  pages={10023--10033},
  year={2020},
}

@inproceedings{tan-etal-2025-improvement,
    title = "Improvement in Sign Language Translation Using Text {CTC} Alignment",
    author = "Tan, Sihan  and
      Miyazaki, Taro  and
      Khan, Nabeela  and
      Nakadai, Kazuhiro",
    editor = "Rambow, Owen  and
      Wanner, Leo  and
      Apidianaki, Marianna  and
      Al-Khalifa, Hend  and
      Eugenio, Barbara Di  and
      Schockaert, Steven",
    booktitle = "Proceedings of the 31st International Conference on Computational Linguistics",
    month = jan,
    year = "2025",
    address = "Abu Dhabi, UAE",
    publisher = "Association for Computational Linguistics",
    url = "https://aclanthology.org/2025.coling-main.219/",
    note = {Accessed 8 July 2026},
    pages = "3255--3266"
}

@misc{tanzer2024youtubesl25largescaleopendomainmultilingual,
      title={YouTube-SL-25: A Large-Scale, Open-Domain Multilingual Sign Language Parallel Corpus}, 
      author={Garrett Tanzer and Biao Zhang},
      year={2024},
      eprint={2407.11144},
      archivePrefix={arXiv},
      primaryClass={cs.CL},
      url={https://arxiv.org/abs/2407.11144},
      note={Accessed 8 July 2026},
}

@inproceedings{shi2022opendomainsignlanguagetranslation,
  title={Open-Domain Sign Language Translation Learned from Online Video},
  author={Shi, Bowen and Brentari, Diane and Shakhnarovich, Gregory and Livescu, Karen},
  booktitle={Proceedings of the 2022 Conference on Empirical Methods in Natural Language Processing},
  pages={6365--6379},
  address={Abu Dhabi, United Arab Emirates},
  publisher={Association for Computational Linguistics},
  year={2022},
  doi={10.18653/v1/2022.emnlp-main.427},
}

@inproceedings{uthus2023youtubeasllargescaleopendomainamerican,
  title={YouTube-ASL: A Large-Scale, Open-Domain American Sign Language-English Parallel Corpus},
  author={Uthus, Dave and Tanzer, Garrett and Georg, Manfred},
  booktitle={Advances in Neural Information Processing Systems},
  volume={36},
  pages={29029--29047},
  year={2023},
  url={https://proceedings.neurips.cc/paper_files/paper/2023/hash/5c61452daca5f0c260e683b317d13a3f-Abstract-Datasets_and_Benchmarks.html},
}

@inproceedings{
zhang2023sltunet,
title={{SLTUNET}: A Simple Unified Model for Sign Language Translation},
author={Biao Zhang and Mathias M{\"u}ller and Rico Sennrich},
booktitle={The Eleventh International Conference on Learning Representations },
year={2023},
url={https://openreview.net/forum?id=EBS4C77p_5S},
note={Accessed 8 July 2026}
}

@inproceedings{ye-etal-2023-cross,
    title = "Cross-modality Data Augmentation for End-to-End Sign Language Translation",
    author = "Ye, Jinhui  and
      Jiao, Wenxiang  and
      Wang, Xing  and
      Tu, Zhaopeng  and
      Xiong, Hui",
    editor = "Bouamor, Houda  and
      Pino, Juan  and
      Bali, Kalika",
    booktitle = "Findings of the Association for Computational Linguistics: EMNLP 2023",
    month = dec,
    year = "2023",
    address = "Singapore",
    publisher = "Association for Computational Linguistics",
    doi = "10.18653/v1/2023.findings-emnlp.904",
    pages = "13558--13571"
}

@inproceedings{10.5555/3600270.3601510,
author = {Chen, Yutong and Zuo, Ronglai and Wei, Fangyun and Wu, Yu and Liu, Shujie and Mak, Brian},
title = {Two-stream network for sign language recognition and translation},
year = {2022},
isbn = {9781713871088},
publisher = {Curran Associates Inc.},
address = {Red Hook, NY, USA},
booktitle = {Proceedings of the 36th International Conference on Neural Information Processing Systems},
articleno = {1240},
numpages = {14},
location = {New Orleans, LA, USA},
series = {NIPS '22}
}

@inproceedings{Chen2022ASM,
  title={A Simple Multi-Modality Transfer Learning Baseline for Sign Language Translation},
  author={Chen, Yutong and Wei, Fangyun and Sun, Xiao and Wu, Zhirong and Lin, Stephen},
  booktitle={Proceedings of the IEEE/CVF Conference on Computer Vision and Pattern Recognition (CVPR)},
  year={2022},
  pages={5120--5130},
}

@misc{albanie2021bbcoxfordbritishsignlanguage,
      title={BBC-Oxford British Sign Language Dataset}, 
      author={Samuel Albanie and Gül Varol and Liliane Momeni and Hannah Bull and Triantafyllos Afouras and Himel Chowdhury and Neil Fox and Bencie Woll and Rob Cooper and Andrew McParland and Andrew Zisserman},
      year={2021},
      eprint={2111.03635},
      archivePrefix={arXiv},
      primaryClass={cs.CV},
      url={https://arxiv.org/abs/2111.03635},
      note={Accessed 8 July 2026},
}

@InProceedings{Zhou_2023_ICCV,
    author    = {Zhou, Benjia and Chen, Zhigang and Clap\'es, Albert and Wan, Jun and Liang, Yanyan and Escalera, Sergio and Lei, Zhen and Zhang, Du},
    title     = {Gloss-Free Sign Language Translation: Improving from Visual-Language Pretraining},
    booktitle = {Proceedings of the IEEE/CVF International Conference on Computer Vision (ICCV)},
    month     = {October},
    year      = {2023},
    pages     = {20871-20881}
}

@inproceedings{
  wong2024signgpt,
  title={Sign2{GPT}: Leveraging Large Language Models for Gloss-Free Sign Language Translation},
  author={Ryan Wong and Necati Cihan Camgoz and Richard Bowden},
  booktitle={The Twelfth International Conference on Learning Representations},
  year={2024},
  url={https://openreview.net/forum?id=LqaEEs3UxU},
  note={Accessed 8 July 2026}
}

@inproceedings{10.1145/3742886.3756703,
author = {Mercanoglu Sincan, Ozge and Bowden, Richard},
title = {Contrastive Pretraining with Dual Visual Encoders for Gloss-Free Sign Language Translation},
year = {2025},
isbn = {9798400719967},
publisher = {Association for Computing Machinery},
address = {New York, NY, USA},
doi = {10.1145/3742886.3756703},
booktitle = {Adjunct Proceedings of the 25th ACM International Conference on Intelligent Virtual Agents},
articleno = {7},
numpages = {7},
location = {Berlin, Germany},
series = {IVA Adjunct '25}
}

@inproceedings{yin2023gloss,
  title={Gloss attention for gloss-free sign language translation},
  author={Yin, Aoxiong and Zhong, Tianyun and Tang, Li and Jin, Weike and Jin, Tao and Zhao, Zhou},
  booktitle={Proceedings of the IEEE/CVF Conference on Computer Vision and Pattern Recognition},
  pages={2551--2562},
  year={2023}
}

@article{Gong2024LLMsAG,
  title={LLMs are Good Sign Language Translators},
  author={Jia Gong and Lin Geng Foo and Yixuan He and Hossein Rahmani and Jun Liu},
  journal={2024 IEEE/CVF Conference on Computer Vision and Pattern Recognition (CVPR)},
  year={2024},
  pages={18362-18372},
}

@inproceedings{Chen2024FactorizedLA,
  title={Factorized Learning Assisted with Large Language Model for Gloss-free Sign Language Translation},
  author={Chen, Zhigang and Zhou, Benjia and Li, Jun and Wan, Jun and Lei, Zhen and Jiang, Ning and Lu, Quan and Zhao, Guoqing},
  booktitle={Proceedings of the 2024 Joint International Conference on Computational Linguistics, Language Resources and Evaluation (LREC-COLING 2024)},
  pages={7071--7081},
  address={Torino, Italia},
  publisher={ELRA and ICCL},
  year={2024},
  url={https://aclanthology.org/2024.lrec-main.620/},
}

@inproceedings{hwang-etal-2025-efficient,
    title = "An Efficient Gloss-Free Sign Language Translation Using Spatial Configurations and Motion Dynamics with {LLM}s",
    author = "Hwang, Eui Jun  and
      Cho, Sukmin  and
      Lee, Junmyeong  and
      Park, Jong C.",
    editor = "Chiruzzo, Luis  and
      Ritter, Alan  and
      Wang, Lu",
    booktitle = "Proceedings of the 2025 Conference of the Nations of the Americas Chapter of the Association for Computational Linguistics: Human Language Technologies (Volume 1: Long Papers)",
    month = apr,
    year = "2025",
    address = "Albuquerque, New Mexico",
    publisher = "Association for Computational Linguistics",
    doi = "10.18653/v1/2025.naacl-long.197",
    pages = "3901--3920",
    ISBN = "979-8-89176-189-6"
}

@article{Jang2025LostIT,
  title={Lost in Translation, Found in Embeddings: Sign Language Translation and Alignment},
  author={Youngjoon Jang and Liliane Momeni and Zifan Jiang and Joon Son Chung and G{\"u}l Varol and Andrew Zisserman},
  journal={ArXiv},
  year={2025},
  volume={abs/2512.08040},
}

@inproceedings{
guo2025bridging,
title={Bridging Sign and Spoken Languages: Pseudo Gloss Generation for Sign Language Translation},
author={Jianyuan Guo and Peike Li and Trevor Cohn},
booktitle={The Thirty-ninth Annual Conference on Neural Information Processing Systems},
year={2025},
url={https://openreview.net/forum?id=p6Huickfj7},
note={Accessed 8 July 2026}
}

@inproceedings{gueuwou-etal-2025-signmusketeers,
    title = "{S}ign{M}usketeers: An Efficient Multi-Stream Approach for Sign Language Translation at Scale",
    author = "Gueuwou, Shester  and
      Du, Xiaodan  and
      Shakhnarovich, Greg  and
      Livescu, Karen",
    editor = "Che, Wanxiang  and
      Nabende, Joyce  and
      Shutova, Ekaterina  and
      Pilehvar, Mohammad Taher",
    booktitle = "Findings of the Association for Computational Linguistics: ACL 2025",
    month = jul,
    year = "2025",
    address = "Vienna, Austria",
    publisher = "Association for Computational Linguistics",
    doi = "10.18653/v1/2025.findings-acl.1157",
    pages = "22506--22521",
    ISBN = "979-8-89176-256-5"
}

@inproceedings{Wu2025SignMouthLM,
  title={Mouthing-Enhanced Multimodal Hierarchical Contrastive Learning for Gloss-Free Sign Language Translation},
  author={Wu, Wenfang and Yuan, Tingting and Li, Yupeng and Wang, Daling and Fu, Xiaoming},
  booktitle={2026 IEEE International Conference on Acoustics, Speech and Signal Processing (ICASSP)},
  pages={8787--8791},
  year={2026},
  doi={10.1109/ICASSP55912.2026.11463378},
}

@inproceedings{Yao2023SignLT,
  title={Sign Language Translation with Iterative Prototype},
  author={Yao, Huijie and Zhou, Wengang and Feng, Hao and Hu, Hezhen and Zhou, Hao and Li, Houqiang},
  booktitle={Proceedings of the IEEE/CVF International Conference on Computer Vision (ICCV)},
  year={2023},
  pages={15592--15601},
}

@INPROCEEDINGS{8575275,
  author={Ahn, Namhyuk and Kang, Byungkon and Sohn, Kyung-Ah},
  booktitle={2018 IEEE/CVF Conference on Computer Vision and Pattern Recognition Workshops (CVPRW)}, 
  title={Image Super-Resolution via Progressive Cascading Residual Network}, 
  year={2018},
  volume={},
  number={},
  pages={904-9048},
  doi={10.1109/CVPRW.2018.00123}}

@inproceedings{ren2019progressiveimagederainingnetworks,
  title={Progressive Image Deraining Networks: A Better and Simpler Baseline},
  author={Ren, Dongwei and Zuo, Wangmeng and Hu, Qinghua and Zhu, Pengfei and Meng, Deyu},
  booktitle={Proceedings of the IEEE/CVF Conference on Computer Vision and Pattern Recognition (CVPR)},
  pages={3937--3946},
  year={2019},
}

@inproceedings{gregor2015drawrecurrentneuralnetwork,
  title={DRAW: A Recurrent Neural Network for Image Generation},
  author={Gregor, Karol and Danihelka, Ivo and Graves, Alex and Rezende, Danilo Jimenez and Wierstra, Daan},
  booktitle={Proceedings of the 32nd International Conference on Machine Learning},
  series={Proceedings of Machine Learning Research},
  volume={37},
  pages={1462--1471},
  year={2015},
  url={https://proceedings.mlr.press/v37/gregor15.html},
}

@inproceedings{Peng2020DeepSF,
  title={Deep Snake for Real-Time Instance Segmentation},
  author={Peng, Sida and Jiang, Wen and Pi, Huaijin and Li, Xiuli and Bao, Hujun and Zhou, Xiaowei},
  booktitle={Proceedings of the IEEE/CVF Conference on Computer Vision and Pattern Recognition (CVPR)},
  year={2020},
  pages={8533--8542},
}

@inproceedings{Ling2019FastIO,
  title={Fast Interactive Object Annotation With Curve-GCN},
  author={Ling, Huan and Gao, Jun and Kar, Amlan and Chen, Wenzheng and Fidler, Sanja},
  booktitle={Proceedings of the IEEE/CVF Conference on Computer Vision and Pattern Recognition (CVPR)},
  year={2019},
  pages={5257--5266},
}

@INPROCEEDINGS{9423326,
  author={Liu, Zichen and Liew, Jun Hao and Chen, Xiangyu and Feng, Jiashi},
  booktitle={2021 IEEE Winter Conference on Applications of Computer Vision (WACV)}, 
  title={DANCE : A Deep Attentive Contour Model for Efficient Instance Segmentation}, 
  year={2021},
  volume={},
  number={},
  pages={345-354},
  doi={10.1109/WACV48630.2021.00039}}

@inproceedings{Tao2022E2ECAE,
  title={E2EC: An End-to-End Contour-based Method for High-Quality High-Speed Instance Segmentation},
  author={Zhang, Tao and Wei, Shiqing and Ji, Shunping},
  booktitle={Proceedings of the IEEE/CVF Conference on Computer Vision and Pattern Recognition (CVPR)},
  year={2022},
  pages={4443--4452},
}

@inproceedings{
madaan2023selfrefine,
title={Self-Refine: Iterative Refinement with Self-Feedback},
author={Aman Madaan and Niket Tandon and Prakhar Gupta and Skyler Hallinan and Luyu Gao and Sarah Wiegreffe and Uri Alon and Nouha Dziri and Shrimai Prabhumoye and Yiming Yang and Shashank Gupta and Bodhisattwa Prasad Majumder and Katherine Hermann and Sean Welleck and Amir Yazdanbakhsh and Peter Clark},
booktitle={Thirty-seventh Conference on Neural Information Processing Systems},
year={2023},
url={https://openreview.net/forum?id=S37hOerQLB},
note={Accessed 8 July 2026}
}

@inproceedings{Bae2024RelaxedRT,
  title={Relaxed Recursive Transformers: Effective Parameter Sharing with Layer-wise LoRA},
  author={Bae, Sangmin and Fisch, Adam and Harutyunyan, Hrayr and Ji, Ziwei and Kim, Seungyeon and Schuster, Tal},
  booktitle={International Conference on Learning Representations (ICLR)},
  year={2025},
  url={https://proceedings.iclr.cc/paper_files/paper/2025/hash/54d6a55225cebbdc16fbb0e45c5bdf2b-Abstract-Conference.html},
}

@misc{xu2026loopingforwardrecursivetransformers,
      title={Looping Back to Move Forward: Recursive Transformers for Efficient and Flexible Large Multimodal Models}, 
      author={Ruihan Xu and Yuting Gao and Lan Wang and Jianing Li and Weihao Chen and Qingpei Guo and Ming Yang and Shiliang Zhang},
      year={2026},
      eprint={2602.09080},
      archivePrefix={arXiv},
      primaryClass={cs.LG},
      url={https://arxiv.org/abs/2602.09080},
      note={Accessed 8 July 2026},
}

@ARTICLE{9887996,
  author={Saharia, Chitwan and Ho, Jonathan and Chan, William and Salimans, Tim and Fleet, David J. and Norouzi, Mohammad},
  journal={IEEE Transactions on Pattern Analysis and Machine Intelligence}, 
  title={Image Super-Resolution via Iterative Refinement}, 
  year={2023},
  volume={45},
  number={4},
  pages={4713-4726},
  doi={10.1109/TPAMI.2022.3204461}}

@INPROCEEDINGS{9711424,
  author={Alaluf, Yuval and Patashnik, Or and Cohen-Or, Daniel},
  booktitle={2021 IEEE/CVF International Conference on Computer Vision (ICCV)}, 
  title={ReStyle: A Residual-Based StyleGAN Encoder via Iterative Refinement}, 
  year={2021},
  volume={},
  number={},
  pages={6711--6720},
  doi={10.1109/ICCV48922.2021.00664}}

@inproceedings{10.1007/978-3-031-20053-3_42,
author = {Shen, Zhiqiang and Liu, Zechun and Xing, Eric},
title = {Sliced Recursive Transformer},
year = {2022},
isbn = {978-3-031-20052-6},
publisher = {Springer-Verlag},
address = {Berlin, Heidelberg},
doi = {10.1007/978-3-031-20053-3_42},
booktitle = {Computer Vision – ECCV 2022: 17th European Conference, Tel Aviv, Israel, October 23–27, 2022, Proceedings, Part XXIV},
pages = {727–744},
numpages = {18},
location = {Tel Aviv, Israel}
}

@inproceedings{Dehghani2018UniversalT,
  title={Universal Transformers},
  author={Dehghani, Mostafa and Gouws, Stephan and Vinyals, Oriol and Uszkoreit, Jakob and Kaiser, Lukasz},
  booktitle={International Conference on Learning Representations (ICLR)},
  year={2019},
  url={https://openreview.net/forum?id=HyzdRiR9Y7},
}

@inproceedings{Elbayad2019DepthAdaptiveT,
  title={Depth-Adaptive Transformer},
  author={Elbayad, Maha and Gu, Jiatao and Grave, Edouard and Auli, Michael},
  booktitle={International Conference on Learning Representations (ICLR)},
  year={2020},
  url={https://openreview.net/forum?id=1KLUy-X5Ayy},
}

@inproceedings{bae2025mixtureofrecursionslearningdynamicrecursive,
  title={Mixture-of-Recursions: Learning Dynamic Recursive Depths for Adaptive Token-Level Computation},
  author={Bae, Sangmin and Kim, Yujin and Bayat, Reza and Kim, Sungnyun and Ha, Jiyoun and Schuster, Tal and Fisch, Adam and Harutyunyan, Hrayr and Ji, Ziwei and Courville, Aaron C. and Yun, Se-Young},
  booktitle={Advances in Neural Information Processing Systems},
  volume={38},
  year={2025},
  url={https://proceedings.neurips.cc/paper_files/paper/2025/hash/8b08bbf8b420faa6eeb4020720582ec7-Abstract-Conference.html},
}

@inproceedings{li-etal-2022-ode,
    title = "{ODE} Transformer: An Ordinary Differential Equation-Inspired Model for Sequence Generation",
    author = "Li, Bei  and
      Du, Quan  and
      Zhou, Tao  and
      Jing, Yi  and
      Zhou, Shuhan  and
      Zeng, Xin  and
      Xiao, Tong  and
      Zhu, JingBo  and
      Liu, Xuebo  and
      Zhang, Min",
    editor = "Muresan, Smaranda  and
      Nakov, Preslav  and
      Villavicencio, Aline",
    booktitle = "Proceedings of the 60th Annual Meeting of the Association for Computational Linguistics (Volume 1: Long Papers)",
    month = may,
    year = "2022",
    address = "Dublin, Ireland",
    publisher = "Association for Computational Linguistics",
    doi = "10.18653/v1/2022.acl-long.571",
    pages = "8335--8351"
}

@inproceedings{
tong2025neural,
title={Neural {ODE} Transformers: Analyzing Internal Dynamics and Adaptive Fine-tuning},
author={Anh Tong and Thanh Nguyen-Tang and Dongeun Lee and Duc Nguyen and Toan Tran and David Leo Wright Hall and Cheongwoong Kang and Jaesik Choi},
booktitle={The Thirteenth International Conference on Learning Representations},
year={2025},
url={https://openreview.net/forum?id=XnDyddPcBT},
note={Accessed 8 July 2026}
}

@inproceedings{10.5555/3327757.3327764,
author = {Chen, Ricky T. Q. and Rubanova, Yulia and Bettencourt, Jesse and Duvenaud, David},
title = {Neural ordinary differential equations},
year = {2018},
publisher = {Curran Associates Inc.},
address = {Red Hook, NY, USA},
booktitle = {Proceedings of the 32nd International Conference on Neural Information Processing Systems},
pages = {6572–6583},
numpages = {12},
location = {Montr\'{e}al, Canada},
series = {NIPS'18}
}

@inproceedings{bai2019deep,
  author    = {Shaojie Bai and J. Zico Kolter and Vladlen Koltun},
  title     = {Deep Equilibrium Models},
  booktitle = {Advances in Neural Information Processing Systems (NeurIPS)},
  year      = {2019},
}

@inproceedings{
zhong2022a,
title={A Neural {ODE} Interpretation of Transformer Layers},
author={Yaofeng Desmond Zhong and Tongtao Zhang and Amit Chakraborty and Biswadip Dey},
booktitle={The Symbiosis of Deep Learning and Differential Equations II},
year={2022},
url={https://openreview.net/forum?id=nA9hvYMQCy},
note={Accessed 8 July 2026}
}

@inproceedings{hu2023self,
  title={Self-Emphasizing Network for Continuous Sign Language Recognition},
  author={Hu, Lianyu and Gao, Liqing and Liu, Zekang and Feng, Wei},
  booktitle={Proceedings of the AAAI Conference on Artificial Intelligence},
  volume={37},
  number={1},
  pages={854--862},
  year={2023},
  doi={10.1609/aaai.v37i1.25164},
}

@inproceedings{ye2024improving,
  title={Improving Gloss-free Sign Language Translation by Reducing Representation Density},
  author={Ye, Jinhui and Wang, Xing and Jiao, Wenxiang and Liang, Junwei and Xiong, Hui},
  booktitle={Advances in Neural Information Processing Systems},
  volume={37},
  year={2024},
  doi={10.52202/079017-3411},
  url={https://proceedings.neurips.cc/paper_files/paper/2024/hash/c225136cfe52a8fd66658bbcf9d894ab-Abstract-Conference.html}
}

@inproceedings{post-2018-call,
  title = "A Call for Clarity in Reporting {BLEU} Scores",
  author = "Post, Matt",
  booktitle = "Proceedings of the Third Conference on Machine Translation: Research Papers",
  month = oct,
  year = "2018",
  address = "Brussels, Belgium",
  publisher = "Association for Computational Linguistics",
  doi = "10.18653/v1/W18-6319",
  pages = "186--191",
}

@InProceedings{Kiziltepe_2025_ICCV,
    author    = {K{\i}z{\i}ltepe, Tu\u{g}\c{c}e and Ta\c{s}y\"urek, S\"umeyye Meryem and Keles, Hacer Yalim},
    title     = {Iterative Latent Refinement for Robust Non-Autoregressive Sign Language Production},
    booktitle = {Proceedings of the IEEE/CVF International Conference on Computer Vision (ICCV) Workshops},
    month     = {October},
    year      = {2025},
    pages     = {4942-4952}
}
\end{document}